\documentclass[conference]{IEEEtran}
\IEEEoverridecommandlockouts 
\usepackage{times}
\usepackage[numbers]{natbib}
\usepackage{multicol}
\usepackage{booktabs}   
\usepackage{tabularx}  
\usepackage{multirow}  
\usepackage{graphicx}  
\usepackage{amsmath}   
\usepackage{amssymb}   
\usepackage{amsfonts}  
\usepackage{xcolor}
\usepackage{makecell}

\usepackage{xcolor}
\usepackage{hyperref}
\usepackage{xcolor} 

\definecolor{cvprblue}{rgb}{0.21, 0.49, 0.74} 
\hypersetup{
    colorlinks=true,      
    linkcolor=cvprblue,   
    citecolor=cvprblue,   
    urlcolor=cvprblue,    
}

\usepackage{marvosym} 

\title{Universal Pose Pretraining for Generalizable Vision-Language-Action Policies}

\author{
    Haitao Lin$^{1,2 \ast, \dagger}$ \quad 
    Hanyang Yu$^{3 \ast, \ddagger}$ \quad 
    Jingshun Huang$^{4,5 \ast, \ddagger}$ \quad 
    He Zhang$^{1,2}$ \quad
    Yonggen Ling$^{1,2}$ \\
    Ping Tan$^{3 \dagger}$ \quad 
    Xiangyang Xue$^{4}$ \quad 
    Yanwei Fu$^{4,5 \dagger}$ 
    \\[0.4em]
    $^1$ Tencent Robotics X \quad $^2$ Futian Laboratory \\
    $^3$ The Hong Kong University of Science and Technology \\
    $^4$ Fudan University \quad $^5$ Shanghai Innovation Institute \\
    \textbf{Project Page:} \href{https://hetolin.github.io/PoseVLA}{\texttt{https://hetolin.github.io/PoseVLA}}
    \thanks{$^\ast$ Equal contribution.}
    \thanks{$^\dagger$ Corresponding authors.}
    \thanks{$^\ddagger$ Work done during an internship at Tencent Robotics X.}
}

\begin{document}
\maketitle

\begin{abstract}

Existing Vision-Language-Action (VLA) models often suffer from feature collapse and low training efficiency because they entangle high-level perception with sparse, embodiment-specific action supervision. Since these models typically rely on VLM backbones optimized for Visual Question Answering (VQA), they excel at semantic identification but often overlook subtle 3D state variations that dictate distinct action patterns. To resolve these misalignments, we propose Pose-VLA, a decoupled paradigm that separates VLA training into a pre-training phase for extracting universal 3D spatial priors in a unified camera-centric space, and a post-training phase for efficient embodiment alignment within robot-specific action space. By introducing discrete pose tokens as a universal representation, Pose-VLA seamlessly integrates spatial grounding from diverse 3D datasets with geometry-level trajectories from robotic demonstrations. Our framework follows a two-stage pre-training pipeline, establishing fundamental spatial grounding via poses followed by motion alignment through trajectory supervision.
Extensive evaluations demonstrate that Pose-VLA achieves state-of-the-art results on RoboTwin 2.0 with a 79.5\% average success rate and competitive performance on LIBERO at 96.0\%. Real-world experiments further showcase robust generalization across diverse objects using only 100 demonstrations per task, validating the efficiency of our pre-training paradigm.

\end{abstract}

\section{Introduction}
The pursuit of general-purpose embodied intelligence has been significantly accelerated by the emergence of Vision-Language-Action (VLA) models. State-of-the-art frameworks such as $\pi$~\cite{black2024pi0, black2025pi0.5} and the GR00T series~\cite{bjorck2025gr00t} have demonstrated remarkable potential in end-to-end robotic manipulation through the adoption of dual-system architectures. In this paradigm, a Vision-Language Model (VLM) acts as a high-level semantic interpreter, while a specialized action expert that employs generative techniques like flow matching~\cite{lipman2022flow} serves as the low-level controller for precise action denoising.

\begin{figure}
    \centering
    \includegraphics[width=1.0\linewidth]{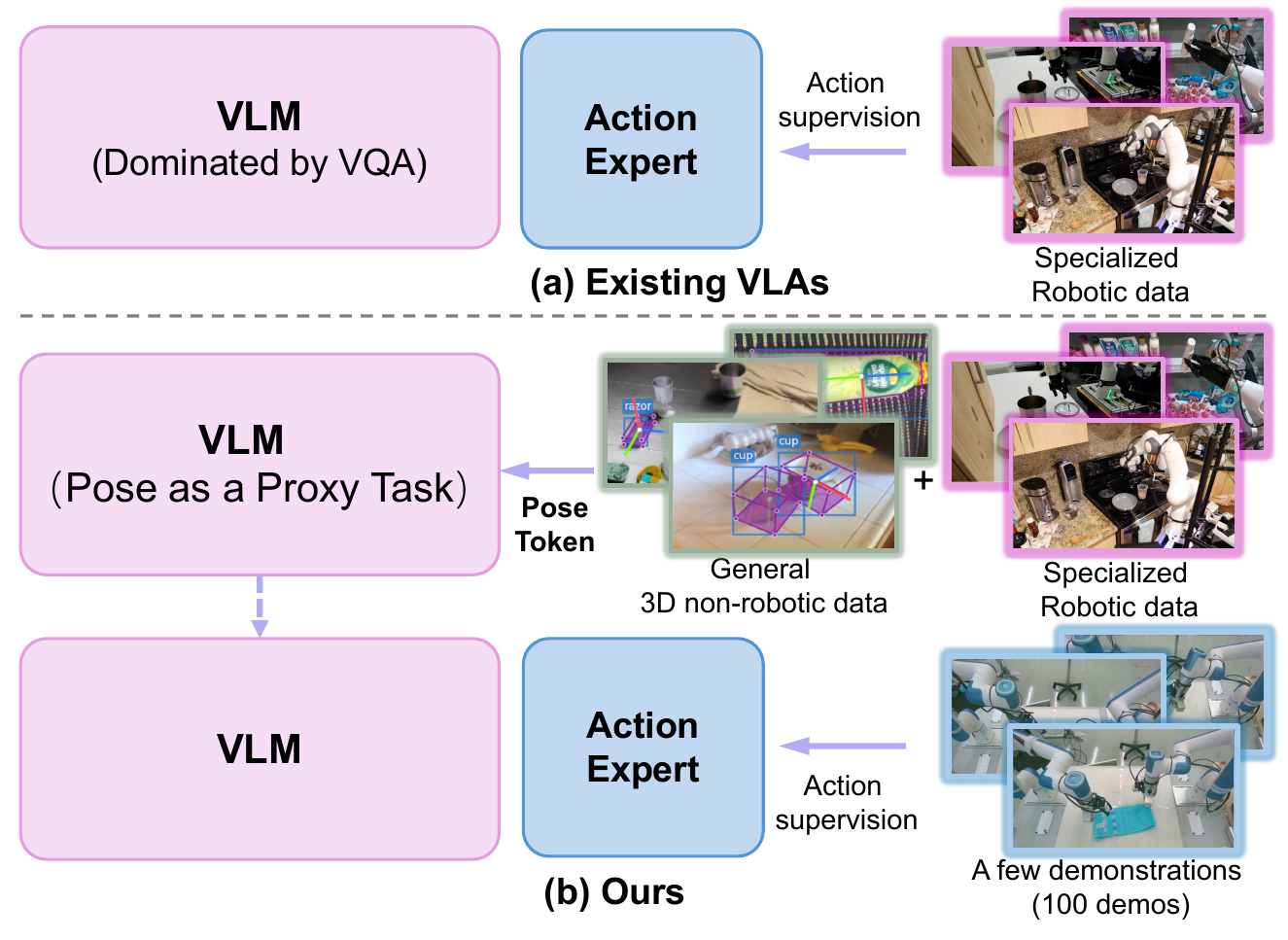}
       \vspace{-0.2in}
    \caption{Overview of Pose-VLA. Unlike previous VLAs that rely solely on sparse action supervision, our approach decouples policy learning into two stages by \textit{using unified pose prediction as a proxy task}: (1) Pre-training, extracting universal 3D spatial priors in a unified camera-centric space; and (2) Alignment, adapting these priors to specific embodiments. This decoupling allows the model to leverage diverse 3D datasets, enabling efficient transfer as backbone when adapting to robotic control with only few-shot fine-tuning.\label{fig:teaser} }
   \vspace{-0.1in}
\end{figure} 

Despite these advances, scaling vision-language representations into physically grounded, generalizable action policies remains a fundamental challenge. While existing VLAs leverage off-the-shelf VLMs~\cite{beyer2024paligemma,bai2025qwen2,qwen3vl2025} as perception backbones, their general semantic features often fail to translate into robust downstream policy performance. This discrepancy is reinforced by recent findings~\cite{zhang2026vlm4vla} showing that fine-tuning on auxiliary tasks like depth estimation or VQA does not inherently guarantee improved control. This persistent gap underscores a critical open question: \textit{how can we efficiently adapt VLMs to acquire transferable embodied priors that directly facilitate downstream policy learning?}

We observe that this failure is not incidental, but structural. It stems from two persistent misalignments between how VLMs are pretrained and how robotic actions are defined.
(1) \textbf{Granularity Mismatch}: VLM pretraining is dominated by tasks such as image–text alignment and visual question answering, which emphasize categorical recognition and high-level semantics. Robotic manipulation, in contrast, depends critically on fine-grained 3D state variations, such as subtle changes in pose, contact geometry, or relative motion that demand qualitatively different actions. As a result, a VLM may correctly recognize what an object is while remaining insensitive to how its physical state evolves. 
(2) \textbf{Data Heterogeneity Gap}: Internet-scale visual corpora lack physical action grounding, while robotic demonstration datasets are scarce, narrowly distributed, and expensive to collect. Existing VLAs struggle to reconcile these two extremes, preventing them from absorbing diverse spatial experiences while remaining relevant to real-world control.

To address these challenges, we propose a decoupled learning paradigm for Vision–Language–Action modeling. Rather than entangling perception with embodiment-specific actions as in prior VLAs~\cite{black2024pi0,kim2024openvla}, we separate learning into two stages as in Fig.~\ref{fig:teaser}:
(1) large-scale pretraining of universal spatial priors in a unified, camera-centric observation space, and
(2) lightweight post-training for embodiment alignment.
This decoupling fundamentally reshapes the objective of large-scale pre-training. By incorporating 3D spatial supervision, the model learns to distinguish subtle 3D variations that dictate distinct action patterns. Once these robust spatial priors are established, the VLM serves as a powerful initialization that makes adapting to a new robotic policy significantly more efficient than learning from scratch.

Building on this principle, we introduce Pose-VLA, a framework that re-centers VLA learning around a universal pose representation. In this architecture, pose acts as a \textit{structural bridge} that maps high-level visual perception to low-level physical execution. Specifically, (1) to bridge the \textit{granularity mismatch}, we represent object states and actions as 3D poses and temporal sequences, explicitly constraining the model to reason over fine-grained spatial supervision. (2) To bridge the \textit{data heterogeneity gap}, pose tokens provide a common language across diverse 3D sources, enabling Pose-VLA to ingest large-scale non-robotic 3D datasets alongside limited robot demonstrations. By unifying these elements, Pose-VLA transforms the VLM into a geometrically-aware backbone optimized for downstream manipulation.

Technically, Pose-VLA builds upon the PaliGemma~\cite{beyer2024paligemma} architecture to unify semantic understanding with spatial grounding. We integrate RGB images with depth maps and camera intrinsics encoded as raymaps to instill intrinsic 3D awareness. A modality masking strategy is employed during training to ensure robustness to varying sensor availability at inference time. We further extend the language model vocabulary with discrete pose tokens that represent 3D transformations in the camera frame, enabling a unified tokenization of spatial information across heterogeneous datasets.

Our training follows a two-stage pipeline. First, spatial foundation pre-training on large-scale 3D datasets establishes geometric grounding. Second, pose alignment pre-training leverages dense multi-view supervision to project robotic trajectories into the same camera-centric observation space. Together, these stages transform the VLM from a semantic describer into a foundation for embodied control. During post-training, we follow~\cite{black2024pi0} by appending a lightweight action expert that maps pretrained representations into robot-specific commands. This strategy ensures that the VLM acquires rich, 3D-aware features that drastically reduce the amount of demonstration data required for downstream policy adaptation.

Our contributions are summarized as follows:
(1) we propose a unified VLM framework that integrates RGB images, depth maps, and camera intrinsics to instill intrinsic 3D awareness, facilitating the effective transfer of vision-language knowledge to robotic control; 
(2) we introduce discrete Pose Tokens as a universal interface (within consistent camera-centric observation space) for aligning and ingesting spatial priors from heterogeneous non-robotic 3D data and specialized robotic demonstrations; and
(3) we curate a comprehensive pre-training corpus comprising 1.4M images with 6.5M 3D annotations for spatial grounding, complemented by approximately 1.55M diverse robotic trajectories for motion alignment.
(4) we demonstrate that Pose-VLA achieves state-of-the-art performance on RoboTwin 2.0 (avg. 79.5\% success) and competitive results on LIBERO (avg. 96.0\%); notably, a single multi-task model exhibits robust real-world generalization across rigid, articulated, and deformable objects, requiring as few as 100 demonstrations per task.

\section{Related Work}
\noindent\textbf{Vision-Language Models for 3D Grounding}
Growing efforts extend VLMs to 3D understanding \cite{guo2025seed1, qwen3vl2025, team2025gemini, yang2025visual, cho2024language, xu2024vlm,team2026hy}, as spatial grounding is crucial for robotic manipulation~\cite{lin2022sar,lin2023pourit,wen2024foundationpose, huang2025cap,zhang2024omni6dpose, lin2022know, wen2022catgrasp}. Unlike earlier language conditioned pipelines that sequentially detect and estimate poses~\cite{sun2023language,fu2024lanpose,cheang2022learning}, VLMs provide the distinct advantage of direct, open vocabulary 3D grounding.
Several methods focus specifically on bridging the gap between vision-language reasoning and 3D geometric tasks. For instance, SpatialLM \cite{mao2025spatiallm} performs 3D grounding from point clouds to generate 3D bounding boxes, although it is limited to a small number of object categories. Spatial-Reasoner \cite{ma2025spatialreasoner} estimates object positions and orientations as intermediate outputs to enhance spatial awareness, while the concurrent N3D-VLA \cite{wang2025n3d} utilizes a massive depth-lifted dataset to enable 3D localization via bounding boxes, yet lacks explicit orientation modeling. 
Unlike previous methods, Pose-VLA utilizes pose tokens to unify object poses and motion trajectories within a shared representation. This enables joint pre-training on heterogeneous 3D datasets and robotic demonstrations, while incorporating camera rays and depth as spatial priors to provide the additional geometric awareness for 3D grounding and robotic manipulation.


\noindent\textbf{Vision-Language-Action Models}
Vision-Language-Action (VLA) models extend the reasoning capabilities of Vision-Language Models (VLMs) to the direct prediction of robotic actions. Current VLA methods can be generally categorized into discrete and continuous action outputs. Discrete approaches ~\cite{rt2, kim2024openvla, pertsch2025fast, zawalski2024robotic, vqvla, liang2025discrete, liu2025hybridvla, vla0} treat action prediction as a next-token prediction task in order to align with the native training paradigm of Large Language Models (LLMs). Alternatively, continuous approaches~\cite{team2024octo, black2024pi0, bjorck2025gr00t, black2025pi0.5, lee2025molmoact, liu2024rdt, shi2025memoryvla, liu2025hybridvla, yang2025instructvla, qu2025eo} add generative heads like diffusion or flow-matching to achieve higher control fidelity.
Despite these advances, how to effectively adapt a VLM’s semantic and reasoning abilities to the action modality remains an open question. Recent studies indicate that while VLM pre-training is fundamental, a model's general VQA competence does not always predict its downstream control performance, highlighting a persistent domain gap. To bridge this gap, strategies like co-training with vision-text data~\cite{black2025pi0.5, Zhou2025chatvla, Lin2025onetwovla, qu2025eo} or incorporating Embodied Chain-of-Thought (ECoT)~\cite{zawalski2024robotic, lee2025molmoact, ecot-light, sun2025emma} have been proposed to preserve reasoning traces. Furthermore, some VLAs~\cite{li2025spatial, kachaev2025don, zhangrepresentation} introduce regularization within the VLM's latent space, such as aligning mid-layer features with a vision teacher or utilizing hidden space world modeling.
While VLM4VLA~\cite{zhang2026vlm4vla} indicates that fine-tuning VLMs on auxiliary tasks rarely translates to better policy performance, we show that 3D pose pre-training establishes a vital geometric foundation for downstream control.

\begin{figure*}
    \centering
    \includegraphics[width=1.0\linewidth]{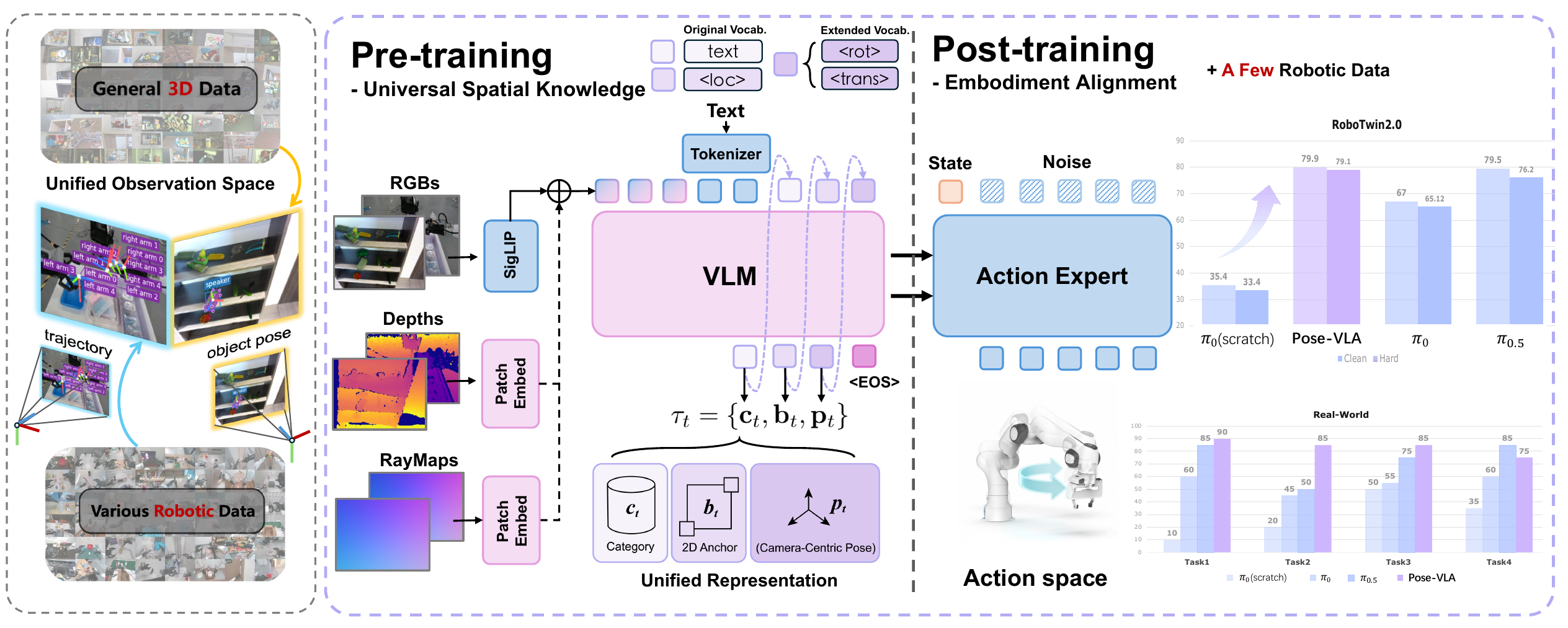}
       \vspace{-0.15in}
    \caption{Pipeline of Pose-VLA. Pose-VLA decouples VLA training into: (1) \textbf{Pre-training} for extracting universal 3D spatial priors in a camera-centric space, and (2) \textbf{Post-training} for embodiment alignment. The VLM predicts a structured sequence $\mathcal{S} = (\tau_1, \dots, \tau_T)$ via next-token prediction, where each tuple $\tau_t = \{\mathbf{c}_t, \mathbf{b}_t, \mathbf{p}_t\}$ consists of a category $\mathbf{c}_t$, 2D box center $\mathbf{b}_t$, and camera-centric pose $\mathbf{p}_t$. To enhance spatial reasoning, auxiliary 3D geometry priors are integrated via additive fusion with RGB embeddings, analogous to positional encodings. This unified format enables seamless knowledge transfer from diverse 3D datasets to robotic domains, achieving robust alignment with minimal demonstrations.
\label{fig:pipeline} }
   \vspace{-0.1in}
\end{figure*}

\section{Method}
\label{sec:method}

\subsection{Preliminaries}
\label{sec:preliminaries}

Given a multi-modal observation $\mathbf{O}$ and a language instruction $\mathbf{L}$, Pose-VLA aims to model the joint probability of an output token sequence $\mathbf{S}$. Following the auto-regressive paradigm, the generation process is formulated as:

\begin{equation}
    P(\mathbf{S} | \mathbf{O}, \mathbf{L}) = \prod_{k=1}^{K} P(\mathbf{s}_k | \mathbf{O}, \mathbf{L}, \mathbf{s}_{<k})
\end{equation}

where $\mathbf{s}_k$ denotes the $k$-th token in a sequence of length $K$, and $\mathbf{L}$ represents the natural language instruction. To bridge high-level semantics with fine-grained 3D control, we define the output as a structured sequence of tuples $\mathcal{S} = (\tau_1, \tau_2, \dots, \tau_T)$, where each tuple $\tau_t$ serves as the fundamental unit for spatial representation:

\begin{equation}
    \tau_t = \{\mathbf{c}_t, \mathbf{b}_t, \mathbf{p}_t\}
\end{equation}

In this formulation, $\mathbf{c}_t$ represents the object category, $\mathbf{b}_t \in \mathbb{R}^2$ denotes the box center in image coordinates, and $\mathbf{p}_t \in SE(3)$ signifies the pose in the camera-centric frame. This design enables $\tau_t$ to serve as a universal geometric primitive: in static contexts, it localizes the object for spatial grounding; in temporal sequences, it characterizes the trajectory waypoints for motion estimation. By unifying 2D centroids with 3D poses, $\tau_t$ provides a versatile representation that describes both discrete physical entities and continuous motion paths within a unified generative framework.

\subsection{VLM architecture}
Our approach adopts PaliGemma~\cite{beyer2024paligemma} as the backbone. PaliGemma employs SigLip~\cite{zhai2023sigmoid} as the visual encoder, which extracts rich semantic features from images. However, PaliGemma is trained mainly on the VQA task and detection task. We empirically find that such supervision on text level or 2D location-level may focus on high-level semantics, while losing the fine-grained details for control-oriented features. Thus, we use the pose as the universal representation to bridge the non-robotics data in computer vision task and robotics data in manipulation task as in Fig.~\ref{fig:pipeline}. As pose is described in the 3D world, it can force the model to understand the spatial relationship in the 3D world. To maintain the inherent 2D visual grounding capabilities, we reuse the  text and \texttt{<loc>} vocabularies from the original PaliGemma model for representing $\mathbf{c}_t$ and $\mathbf{b}_t$.

To incorporate 3D priors, we leverage auxiliary inputs, including depth maps and camera intrinsics, which are commonly available in 3D-annotated, non-robotic datasets. The inclusion of such auxiliary information has been shown to markedly improve 3D prediction performance at test time. 

For 3D-annotated non-robotics data, poses are defined within the camera coordinate frame. To maintain alignment with these diverse geometric datasets, our approach requires the model to predict actions directly within the camera frame of the image view, rather than estimating robot-centric actions defined in the robot's base frame. This design ensures a shared camera-centric representation across heterogeneous data sources, enabling the model to fully leverage the 3D awareness acquired during VLM pre-training for precise robotic control.

\subsection{Unified Pose Representation}

We adopt 3D pose as a universal representation to unify the feature spaces of non-robotic 3D datasets and robotic manipulation data. This choice is motivated by the fact that both objects and robotic grippers can be effectively parameterized within this shared geometric space. By imposing this representation, we constrain the model to estimate spatial coordinates directly from visual observations within the camera frame, which naturally fosters the development of robust 3D spatial grounding capabilities. To implement this, we introduce a dedicated \textbf{pose token} defined in the camera coordinate system. This design choice explicitly addresses the challenge of coordinate misalignment encountered during pre-training, facilitating a seamless integration of diverse data sources into a consistent framework.

We represent object states and robotic motion trajectories using a parameterization of translation and rotation. In the following sections, we detail the implementation of these pose tokens and demonstrate their versatility across both general 3D spatial grounding and embodied robotic manipulation tasks.

\noindent\textbf{Object-Level Representation.}
Each object is characterized by its rigid-body transformation, comprising translation and rotation. We discretize these continuous parameters to integrate them into the VLM as an extended vocabulary of specialized tokens. For rotation, we perform uniform quantization of the Euler angles and assign a dedicated \texttt{<rot>} token to each principal axis. 

Regarding translation, an empirical analysis of our large-scale pre-training statistics reveals a distinct distributional divergence between the lateral ($x, y$) and longitudinal ($z$) axes as shown in our Appendix. While the $x$ and $y$ distributions exhibit comparable means and variances, the $z$-axis (depth) distribution is significantly shifted, reflecting the inherent perspective projection and depth range of the camera. To accommodate this discrepancy and enforce depth-aware geometric grounding, we employ a shared \texttt{<trans\_xy>} token for the $x$ and $y$ dimensions and a distinct \texttt{<trans\_z>} token for the depth dimension.

To evaluate 3D grounding performance in general 3D benchmark~\cite{brazil2023omni3d}, we incorporate object scale into the representation. Since object dimensions across all three axes exhibit similar statistical characteristics, we represent object scale using a unified \texttt{<size>} token for spatial dimensions.

By mapping image observations into structured pose tokens, Pose-VLA establishes a direct correspondence between image pixels and physical entities within a metric-aligned coordinate system. Unlike previous methods \cite{bai2025qwen2,man2025locateanything3d} that rely on decomposing numerical coordinates into individual digit tokens, our tokenization scheme significantly enhances efficiency by reducing the required token sequence length per coordinate. Crucially, this approach ensures that all task data are optimized within a shared real-world metric space, enabling a unified physical vocabulary across both perception and action. By supervising the model to estimate target object poses, we drive the network to implicitly learn 3D spatial features, leading to a level of fine-grained geometric reasoning that far surpasses the capabilities of conventional VLMs dominated by VQA supervision.

\noindent\textbf{Trajectory-Level Representation.}
Robot actions can be parameterized by the 6-DoF pose of the gripper. By adopting this convention, the gripper pose can directly reuse the discretized tokens established for object-level representations. For dynamic manipulation, we represent action trajectories as temporally ordered sequences of these unified pose tokens.

Unlike contemporary VLA frameworks~\cite{kim2024openvla} that define actions as joint angles or base-frame poses, our approach addresses the fundamental misalignment between the observation space (camera-centric) and the action space (robot-centric). Such spatial discrepancies often hinder cross-embodiment generalization, as the model must implicitly bridge the gap between localized visual observations and a global robot-base coordinate system. To resolve this conflict, we project all action trajectories into the respective camera coordinate frame of each view. This alignment ensures that both non-robotic 3D grounding data and robotic demonstration data share a consistent geometric reference frame.

By supervising the model to estimate future trajectories directly within the image observation space, we provide a significantly denser and more grounded supervision signal compared to mapping pixels to embodiment-specific tokens~\cite{kim2024openvla}. Furthermore, this camera-centric formulation enhances implicit 3D correspondence learning across diverse viewpoints, such as head-mounted and wrist-mounted cameras. This dense supervision encourages the model to learn representative, control-oriented features that are sensitive to subtle variations between consecutive observations. Ultimately, representing the gripper and objects within a shared token space unifies the optimization objectives of large-scale computer vision and robotics tasks, facilitating the scalable transfer of 3D geometric knowledge to downstream robotic manipulation.

\subsection{Adding Prior Conditioning}

To enhance the framework's geometric awareness, we incorporate auxiliary spatial priors. Inspired by Pow3R \cite{jang2025pow3r}, these embeddings are integrated via additive fusion with the RGB token embeddings prior to the initial Transformer block, analogous to standard positional encodings. 

\noindent\textbf{Camera Ray Encoding.} 
To establish a direct correspondence between pixels and their physical viewing directions, we construct camera rays derived from the intrinsic matrix $\mathbf{K} \in \mathbb{R}^{3\times 3}$. The ray $\mathbf{r}$ for each pixel $(u, v)$ is computed as $\mathbf{r} = \mathbf{K}^{-1} [u, v, 1]^\top$, which encodes the viewing direction relative to the camera's optical center. This ray encoding enables the model to effectively process non-centered crops and facilitates high-resolution inference by providing absolute geometric anchors. Following the RGB processing pipeline, these dense ray maps are patchified and projected into the latent space before being fused with the visual features.

\noindent\textbf{Depth Prior Integration.} 
For the depth prior, given a depth map $\mathbf{D}$ and its corresponding sparsity mask $\mathbf{M}$, we deliberately avoid value normalization to maintain explicit alignment with the metric space. We stack the depth and mask into a joint representation $[\mathbf{D}, \mathbf{M}] \in \mathbb{R}^{H \times W \times 2}$, which is then patchified and embedded. This configuration allows the encoder to robustly handle varying levels of sparsity through integrated valid-pixel masking. 

This multi-modal conditioning ensures the seamless integration of semantic and 3D geometric information, significantly bolstering the model's capacity for precise 3D reasoning in complex manipulation scenarios.

\subsection{Training Strategy}

\noindent\textbf{Pre-training stage.} 
In this stage, we employ a next-token prediction loss to instill fundamental spatial priors into the model. To preserve the extensive vision-language capabilities of the base VLM, we jointly supervise the generation of 2D localization tokens and our extended 3D pose tokens. Given the multi-modal input $\mathbf{O}$ and language instruction $\mathbf{L}$, we utilize the standard auto-regressive objective:

\begin{equation}
\mathcal{L}_{pre}(\theta) = - \sum_{k=1}^{K} \log p_{\theta}(\mathbf{s}_k | \phi(\mathbf{O}), \mathbf{L}, \mathbf{s}_{<k})
\end{equation}

where $s_k$ denotes the $k$-th token in the structured sequence $\mathcal{S}$, and $\phi(\mathbf{O})$ represents the modality masking strategy. Specifically, during training, we randomly zero-pad raymaps or depth values to ensure the model maintains robust inference capabilities even with RGB-only inputs. Furthermore, for depth images, we sample sparse depth points to enhance robustness to diverse sensor noise patterns. This training phase equips the VLM with coarse yet semantically grounded spatial awareness, enabling it to transition from a semantic describer to a geometric reasoner.

\noindent\textbf{Post-training stage.} 
For embodiment alignment, we follow~\cite{black2024pi0} by appending a lightweight action expert to map the pre-trained representations from the VLM backbone into robot-specific commands. This decoupling allows our primary focus to remain on large-scale spatial prior acquisition during pre-training. Specifically, we employ a flow matching objective~\cite{lipman2022flow,liu2022rectified} to denoise the action tokens, where the VLM backbone interacts with the action expert through self-attention layers to provide semantic and geometric conditioning. Notably, the action expert is trained from scratch without inheriting initialization parameters from~\cite{black2024pi0}, ensuring that the learned alignment is natively grounded in our Pose-VLA representations.

\noindent\textbf{Remark on Inference:} We emphasize that 3D pose prediction serves strictly as an \textit{auxiliary proxy task} during pre-training to learn spatial representations. At inference time, the action expert maps implicit features directly to continuous actions end-to-end, introducing zero overhead for explicit pose or camera trajectory estimation.

\subsection{Pre-training Datasets}
Our pre-training utilizes two types of datasets:
\noindent\textbf{(1) Spatial Grounding and 3D Perception.}
We curate 1.4M images with 6.5M 3D annotations across three pillars: 
(1) \textit{General 3D Detection} via Omni3D~\cite{brazil2023omni3d} (Objectron, SUN RGB-D, ARKitScene); 
(2) \textit{6D Pose Estimation} via Omni6DPose~\cite{zhang2024omni6dpose} (SOPE, ROPE); and 
(3) \textit{Manipulation-centric Perception} from the BOP Challenge (YCB-V, LineMOD, HOT3D). This mixture of real and synthetic data ensures robust awareness of 3D geometry and lighting variations.
\noindent\textbf{(2) Robotic Trajectory Pre-training.}
This stage aligns spatial features with motion control using nearly 1.55M trajectories. All end-effector poses are transformed into a unified camera-centric frame via calibration parameters provided by datasets.
(1) \textit{AgibotWorld Beta (Real):} 1M trajectories from 100 isomorphic robots, covering 200 tasks and 87 atomic skills. 
(2) \textit{InternData-A1 (Sim):} 550K trajectories across heterogeneous platforms, including ARX, Aloha, Franka. Spanning 4 embodiments and 227 scenes, it provides dense supervision for articulated objects and long-horizon tasks.

\noindent\textbf{Implementation Details.}
The model is optimized via AdamW with a cosine schedule, reaching a peak learning rate of $5 \times 10^{-5}$ after a 1,000-step warmup. This is followed by a cosine decay to a minimum of $2.5 \times 10^{-6}$. Training is performed on 16 NVIDIA H20 GPUs with bfloat16 precision and a per-GPU batch size of 8.
Specifically, both 3D spatial grounding and trajectory estimation tasks are completed within 2 days each on 16 H20 GPUs. This approach significantly reduces GPU hours compared to previous VLA training~\cite{bjorck2025gr00t,black2024pi0} while effectively instilling robust spatial priors for the VLM.

\section{Experiment}
\subsection{Evaluation in 3D Grounding Benchmarks}
\noindent\textbf{Baselines.}
For the 3D grounding task, we compare against recent open-source and closed-source vision-language models capable of end-to-end 3D spatial grounding directly from language prompts. Our open-source baselines include the Qwen3-VL series~\cite{qwen3vl2025} and the Visual Spatial Tuning (VST) models~\cite{yang2025visual}. For closed-source baselines, we evaluate against Seed1.5-VL~\cite{guo2025seed1}, Gemini-2.0-Pro~\cite{comanici2025gemini}, and the state-of-the-art Gemini Robotics-ER~\cite{team2025gemini}, as their settings align most closely with our experimental protocol.

\noindent\textbf{Evaluation Metrics.} 
Following the protocols established in Qwen3-VL \cite{qwen3vl2025} and Seed1.5-VL \cite{guo2025seed1}, we adopt Average Precision (AP) as our primary evaluation metric. Each evaluation input consists of an image-text pair, where the textual prompt specifies the target object category. To ensure a fair and consistent comparison with existing VLMs, we set the Intersection over Union (IoU) threshold to 0.15 and report mAP@0.15 on the Omni3D test set~\cite{brazil2023omni3d}. 
We primarily conduct evaluations on the SUN-RGBD~\cite{song2015sun} and Objectron~\cite{ahmadyan2021objectron} subsets of Omni3D~\cite{brazil2023omni3d}, which cover a wide range of indoor and tabletop environments.
In line with previous benchmarks, the detection confidence for all predictions is fixed at 1.0 to evaluate the spatial grounding accuracy independently of confidence scoring.

\noindent\textbf{Results and Analysis}
Table \ref{tab:spatial_performance} summarizes the 3D spatial grounding performance across various foundational models. Our proposed Pose-VLA achieves a substantial performance leap on the \textit{Objectron} dataset, reaching an $AP_{15}$ of 87.3. This represents a 16.1\% absolute improvement over the strongest open-source baseline, Qwen3-VL~\cite{qwen3vl2025} (235B, \textit{Thinking}), and a significant margin over closed-source models such as Seed1.5-VL~\cite{guo2025seed1} and Gemini-2.0-Pro~\cite{team2025gemini}. On the SUN RGB-D benchmark~\cite{song2015sun}, Pose-VLA delivers a competitive score of 45.5, outperforming all open-source variants and the VST-3B-SFT~\cite{yang2025visual} model, while remaining closely comparable to the state-of-the-art Gemini Robotics-ER~\cite{team2025gemini}. 

The results on Objectron~\cite{ahmadyan2021objectron} are particularly significant as they highlight the model's robust capability in handling diverse daily objects essential for robotic manipulation, such as \textit{bottles, cups, and cameras}. While SUN RGB-D evaluates general indoor scene understanding, Objectron demands precise geometric localization of individual objects. The fact that Pose-VLA maintains superior spatial awareness across both large-scale indoor environments and object-centric contexts validates our \textit{pose-centric} pre-training paradigm. 

\begin{figure*}
    \centering
    \includegraphics[width=0.9\linewidth]{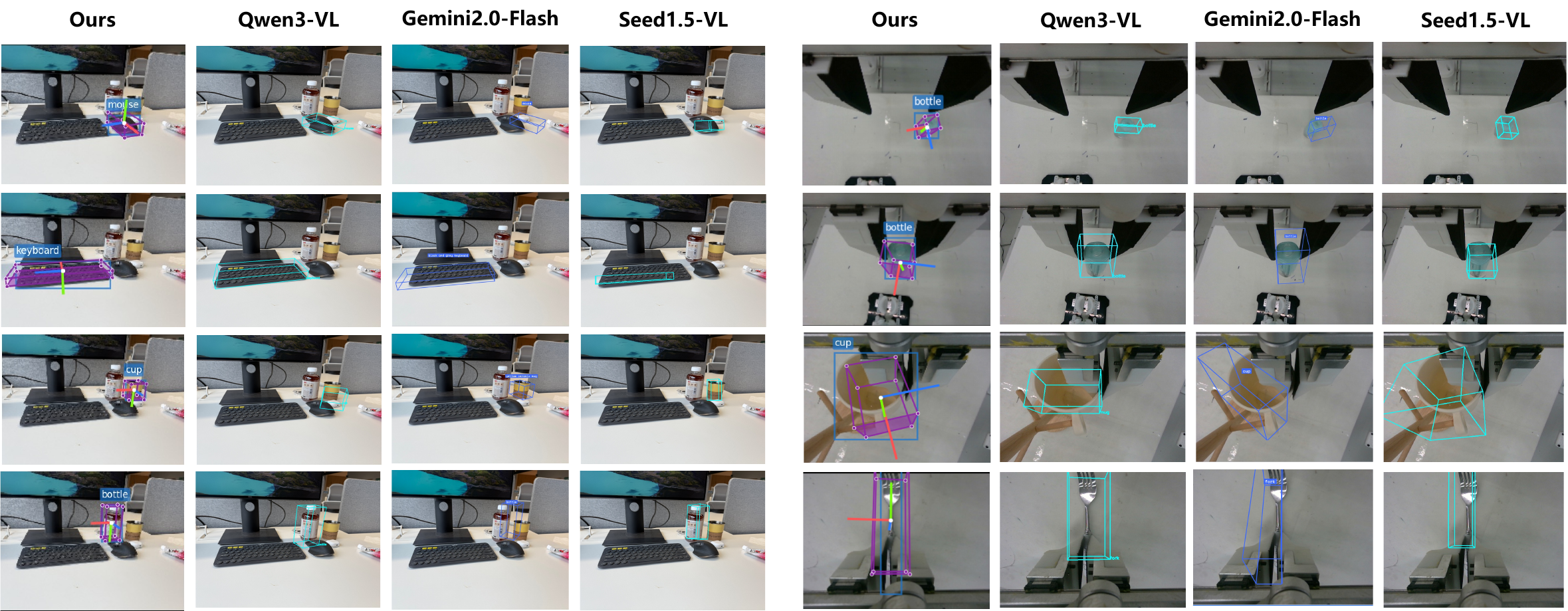}
       \vspace{-0.1in}
    \caption{Generalization of 3D spatial grounding across unseen scenarios. Pose-VLA exhibits robust generalization across various unseen settings, ranging from indoor tabletop layouts to complex robotic manipulation workspaces, providing more precise geometric localization than baseline methods. \label{fig:vis_3d} }
   \vspace{-0.1in}
\end{figure*}

Crucially, Fig.~\ref{fig:vis_3d} demonstrates that Pose-VLA generalizes robustly across unseen scenarios, ranging from tabletop layouts to complex robotic workspaces. Unlike baselines that frequently suffer from orientation misalignment in novel environments, Pose-VLA provides precise localization and orientation estimation. This superior generalization highlights the effectiveness of utilizing auxiliary spatial priors and diverse 3D datasets, enabling the model to maintain high-fidelity spatial awareness even in out-of-distribution scenarios.

\begin{table}[h]
\centering
\renewcommand{\arraystretch}{1.2} 
\setlength{\belowcaptionskip}{5pt}

\caption{Comparison of 3D Spatial Grounding Performance using the $AP_{15}$ metric across different VLM backbones. Baseline results are sourced from the official Qwen3-VL report \cite{qwen3vl2025}.}\label{tab:spatial_performance}
\footnotesize
\resizebox{\columnwidth}{!}{
\begin{tabular}{llcc}
\toprule
Category & Model & SUN RGB-D~\cite{song2015sun} & Objectron~\cite{ahmadyan2021objectron} \\ 
\midrule
\multirow{6}{*}{\shortstack[l]{Open \\ Source}} & Qwen3-VL~\cite{qwen3vl2025} (235B, Thinking) & 34.9 & 71.2 \\
 & Qwen3-VL~\cite{qwen3vl2025} (235B, Instruct) & 39.4 & - \\
 & Qwen3-VL~\cite{qwen3vl2025} (4B, Instruct) & 34.7 & - \\
 & Qwen3-VL~\cite{qwen3vl2025} (2B, Instruct) & 33.8 & - \\
 & VST-SFT(3B)~\cite{yang2025visual} & 37.3 & - \\
 & VST-RL(3B)~\cite{yang2025visual} & 40.1 & - \\
\midrule
\multirow{3}{*}{\shortstack[l]{Closed \\ Source}} & Seed1.5-VL~\cite{guo2025seed1} & 33.5 & 8.1 \\
 
 & Gemini-2.0-Pro~\cite{comanici2025gemini} & 32.5 & 5.5 \\
 & Gemini Robotics-ER~\cite{team2025gemini} & \textbf{48.3} & - \\
\midrule
 & Ours-VLM (3B) & 45.5 & \textbf{87.3} \\
\bottomrule
\end{tabular}
}
\end{table}

\subsection{Evaluation in Simulation Benchmarks}

To evaluate the cross-embodiment generalization and task-level robustness of Pose-VLA, we conduct extensive experiments on two large-scale simulation benchmarks: RoboTwin-2.0 \cite{chen2025robotwin} and LIBERO \cite{liu2024libero}.

\noindent\textbf{RoboTwin-2.0.} 
To evaluate the generalization capabilities of our method, we perform \textit{multi-task training} on the RoboTwin-2.0 benchmark~\cite{chen2025robotwin}. Specifically, Pose-VLA and all baseline models~\cite{black2024pi0, black2025pi0.5, beyer2024paligemma} are trained on a comprehensive dataset comprising 2,500 demonstrations collected in clean scenes (50 per task), augmented with an additional 25,000 demonstrations gathered in heavily randomized scenes (500 per task). To ensure a strictly fair comparison across this dataset, all models are initialized from their respective pre-trained checkpoints and fine-tuned using an identical training budget of 80K optimization steps with a total batch size of 32.


To rigorously assess performance under distribution shifts, we conduct inference across both "Easy" and "Hard" scenarios. While the "Easy" setting serves as a controlled baseline, the "Hard" scenes introduce significant visual and geometric noise to evaluate the model's robustness. We report the average success rate over 100 evaluation trials per task. 

\noindent\textbf{LIBERO.} 
Following the protocol of OpenVLA \cite{kim2024openvla}, we evaluate our model on four distinct task suites: LIBERO-Spatial, LIBERO-Object, LIBERO-Goal, and LIBERO-Long. Each suite comprises 10 diverse tasks with 500 demonstrations in total. These benchmarks allow us to evaluate Pose-VLA across multiple dimensions, ranging from spatial reasoning to long-horizon planning. We compare our approach against a broad range of state-of-the-art generalist policies, including OpenVLA \cite{kim2024openvla}, SpatialVLA \cite{qu2025spatialvla}, $\pi_0$ \cite{black2024pi0}, $\mathbf{\pi}_{0.5}$~\cite{black2025pi0.5}, $\pi_0$-FAST \cite{pertsch2025fast}, CoT-VLA \cite{zhao2025cot}, WorldVLA \cite{cen2025worldvla}, and GR00T-N1 \cite{bjorck2025gr00t}.

\begin{table*}[ht]
  \centering
  \footnotesize
  \setlength{\tabcolsep}{6.5pt}
  \caption{\textbf{Comparison on the RoboTwin 2.0 simulation.} All models are post-trained for 80K steps (batch size=32) for fair comparison, except for the PoseVLA(Upperbound), which is trained for 100K steps (batch size=288, chunk size=16).}
  \label{tab:robotwin-subset}
  \begin{tabular}{*{1}{>{\centering\arraybackslash}m{3.2cm}} *{10}{>{\centering\arraybackslash}m{0.75cm}}}
    \toprule
    \multirow{2}{*}{\textbf{Manipulation Task}} 
      & \multicolumn{2}{c}{$\mathbf{\pi}_0$~\cite{black2024pi0}}
      & \multicolumn{2}{c}{$\mathbf{\pi}_{0.5}$~\cite{black2025pi0.5}}
      & \multicolumn{2}{c}{PaliGemma\_expert}
      & \multicolumn{2}{c}{Pose-VLA(w/o depth)} 
      & \multicolumn{2}{c}{PoseVLA(Upperbound)} \\
    & Easy & Hard & Easy & Hard & Easy & Hard & Easy & Hard & Easy & Hard \\ 
    \midrule
\textit{Adjust Bottle} & 70\% & 57\% & 62\% & 69\% & 67\% & 47\% & 97\% & 77\% & 100\% & 92\% \\
\textit{Beat Block Hammer} & 80\% & 73\% & 78\% & 93\% & 40\% & 37\% & 100\% & 87\% & 80\% & 68\% \\
\textit{Click Alarmclock} & 83\% & 73\% & 92\% & 89\% & 80\% & 57\% & 83\% & 93\% & 98\% & 100\% \\
\textit{Dump Bin Bigbin} & 100\% & 90\% & 89\% & 97\% & 73\% & 77\% & 97\% & 97\% & 98\% & 98\% \\
\textit{Grab Roller} & 93\% & 97\% & 100\% & 100\% & 53\% & 57\% & 97\% & 100\% & 100\% & 100\% \\
\textit{Handover Block} & 33\% & 40\% & 84\% & 57\% & 7\% & 17\% & 73\% & 80\% & 92\% & 86\% \\
\textit{Lift Pot} & 63\% & 73\% & 100\% & 85\% & 37\% & 30\% & 100\% & 97\% & 100\% & 98\% \\
\textit{Move Pillbottle Pad} & 73\% & 67\% & 85\% & 61\% & 10\% & 20\% & 90\% & 87\% & 88\% & 90\% \\
\textit{...(50 tasks)} & ...& ...& ...& ...& ...& ...& ...& ...& ...& ...\\
\textit{Open Laptop} & 73\% & 70\% & 88\% & 96\% & 67\% & 70\% & 93\% & 93\% & 96\% & 96\% \\
\textit{Pick Dual Bottles} & 47\% & 50\% & 21\% & 63\% & 10\% & 0\% & 87\% & 87\% & 98\% & 74\% \\
\textit{Place A2b Left} & 60\% & 43\% & 84\% & 82\% & 27\% & 10\% & 90\% & 80\% & 100\% & 88\% \\
\textit{Place Cans Plasticbox} & 83\% & 73\% & 90\% & 84\% & 23\% & 17\% & 97\% & 87\% & 100\% & 100\% \\
\textit{Place Container Plate} & 93\% & 100\% & 89\% & 95\% & 90\% & 80\% & 97\% & 100\% & 100\% & 100\% \\
\textit{Place Dual Shoes} & 63\% & 50\% & 93\% & 75\% & 3\% & 0\% & 87\% & 77\% & 92\% & 98\% \\
\textit{Place Object Scale} & 70\% & 43\% & 82\% & 80\% & 10\% & 10\% & 67\% & 83\% & 90\% & 84\% \\
\textit{Place Phone Stand} & 70\% & 57\% & 83\% & 81\% & 30\% & 20\% & 87\% & 87\% & 88\% & 94\% \\
\textit{Place Shoe} & 87\% & 80\% & 96\% & 93\% & 23\% & 30\% & 100\% & 87\% & 98\% & 100\% \\
    \midrule
    \textbf{\textit{Average (\%)}} & 67.00 & 65.12  & 79.48 & 76.16 & 35.40 & 33.36 & 79.91 & 79.10 & \textbf{89.40} & \textbf{88.60} \\
    \bottomrule
  \end{tabular}
\end{table*}

\begin{table}[t]
  \centering
  \footnotesize           
  \setlength{\tabcolsep}{4pt}  
  \caption{\textbf{Success rates (\%) on LIBERO benchmark.} \textbf{Bold} and \underline{underline} denote the \textbf{best} and \underline{second best} results across four suites and their \textbf{Average}.}

  \label{tab:liberobaselinessmall}
  \begin{tabular}{lccccc}
    \toprule
    \textbf{Method} & \textbf{Spatial} & \textbf{Object} & \textbf{Goal} & \textbf{Long} & \textbf{Avg} \\
    \midrule
    OpenVLA~\cite{kim2024openvla}      & 84.7 & 88.4 & 79.2 & 53.7 & 76.5  \\
    SpatialVLA~\cite{qu2025spatialvla}    & 88.2 & 89.9 & 78.6 & 55.5 & 78.1 \\
    CoT-VLA~\cite{zhao2025cot}    & 87.5 & 91.6 & 87.6 & 69.0 & 83.9 \\
    WorldVLA~\cite{cen2025worldvla}    & 87.6 & 96.2 & 83.4 & 60.0 & 79.1 \\
    GR00T-N1~\cite{bjorck2025gr00t} & 94.4 & 97.6 & 93.0 & 90.6 & 93.9 \\
    $\mathbf{\pi}_0$~\cite{black2024pi0}       & 96.8 & 98.8 & 95.8 & 85.2 & 94.1  \\
    $\mathbf{\pi}_{0.5}$~\cite{black2025pi0.5}       & \textbf{98.8} & \textbf{98.2} & \textbf{98.0} & \textbf{92.4} & \textbf{96.8}  \\
    $\mathbf{\pi}_0$-FAST~\cite{pertsch2025fast}       & 96.4 & 96.8 & 88.6 & 60.2 & 85.5  \\
    Pose-VLA(w/o depth)       & \underline{96.5} & \underline{98.0} & \underline{97.1} & \textbf{92.4} & \underline{96.0}  \\
    \bottomrule
  \end{tabular}
\end{table}

\noindent\textbf{Results and Analysis.} 
Table \ref{tab:robotwin-subset} and Table \ref{tab:liberobaselinessmall} present the comprehensive evaluation results across simulation benchmarks. 
To ensure a fair comparison with RGB-only baselines, we evaluate Pose-VLA using only RGB input by masking out depth and raymap modalities. 
On RoboTwin 2.0, Pose-VLA establishes a new state-of-the-art, achieving an average success rate of 79.1\% in the challenging \textit{Hard} setting. This represents a substantial margin of 14.0\% over the strong baseline $\pi_0$ and a dramatic improvement of over 45\% compared to the vanilla PaliGemma baseline, confirming that our spatial pre-training effectively prevents feature collapse where standard VQA-oriented backbones fail. Extending this evaluation to LIBERO, Pose-VLA demonstrates superior cross-task generalization with an overall average of 96.0\%, surpassing $\pi_0$ and ranking second only to $\pi_{0.5}$. Notably, in the LIBERO-Long suite which demands multi-stage reasoning, our model attains 92.4\%, tying with $\pi_{0.5}$ for the top position. Thus, these results validate that integrating 3D spatial priors not only enhances robustness against visual perturbations but also provides the structural consistency required for long-horizon planning.

\subsection{Evaluation in Real-world Tasks.}
\noindent\textbf{Real-world Experimental Setup.} 
We evaluate our model using a Dual-arm Xtrainer robotic platform, where each arm is equipped with a 1-DoF parallel gripper. For visual perception, we employ a multi-camera configuration: a RealSense D455 depth camera serves as the head-mounted \textit{Eye-on-Base} sensor for global scene understanding, while a RealSense D405 depth camera is integrated as an \textit{Eye-on-Hand} sensor to provide localized, high-resolution visual feedback.

\begin{figure}[t]
    \centering
    \includegraphics[width=0.85\linewidth]{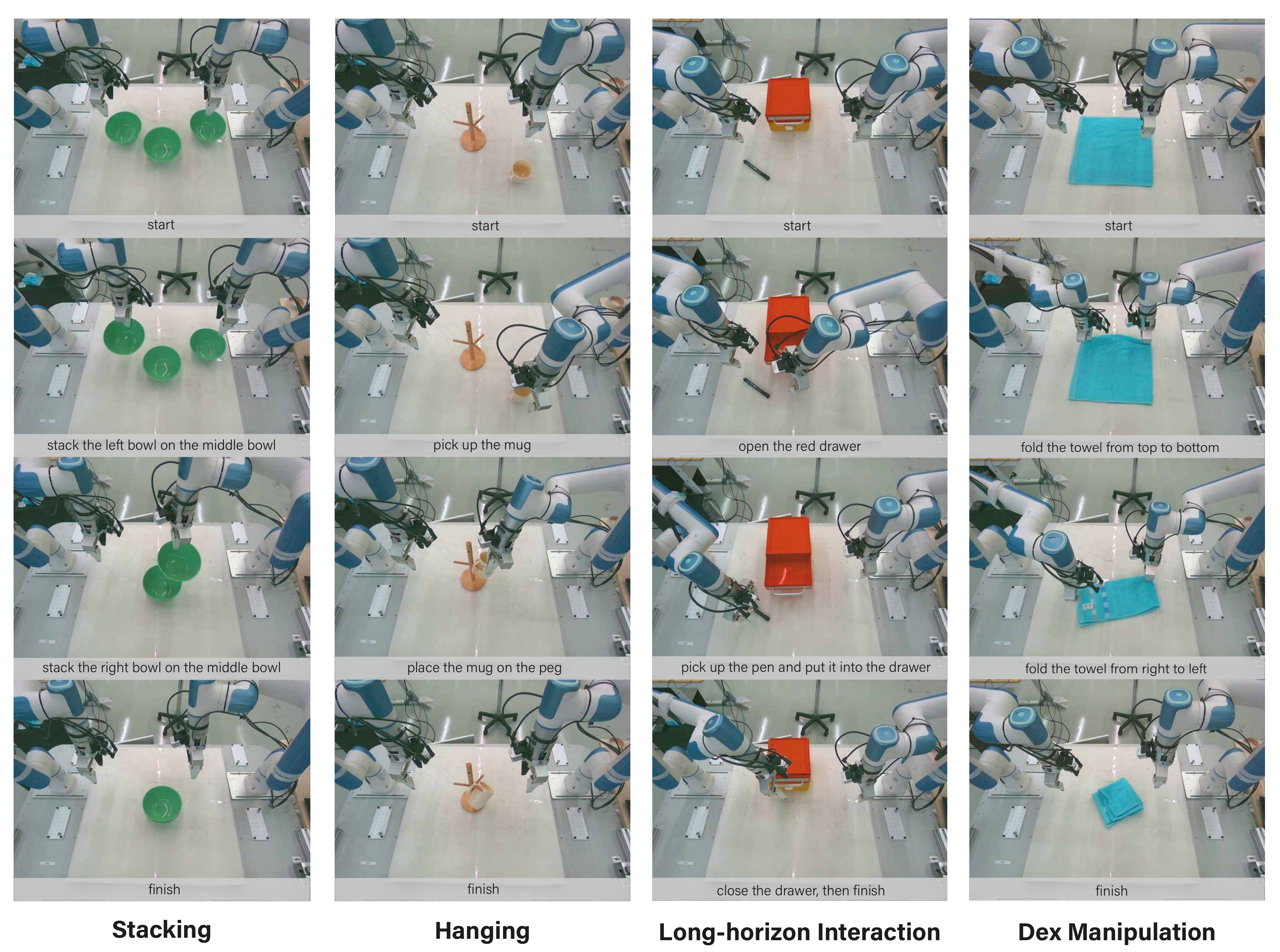}
       \vspace{-0.15in}
    \caption{Real-world setup of four representative tasks. Our platform uses a dual-arm Xtrainer with head and wrist cameras. The benchmark includes: (1)  \textit{Tableware Arrangement}, (2) \textit{Hanging} a mug, (3) \textit{Long-horizon} drawer interaction, and (4) \textit{Deformable} towel folding. Success rates are evaluated over 60 trials per task. \label{fig:real} }
   \vspace{-0.1in}
\end{figure}
\noindent\textbf{Tasks and Evaluation.} 
To verify the generalizability of Pose-VLA, we define four representative and challenging manipulation tasks as in Fig.~\ref{fig:real}: 
(1) \textit{Stacking}: stacking three nested bowls; 
(2) \textit{Hanging}: hanging a mug onto a designated peg on a wooden stand; 
(3) \textit{Long-horizon Interaction}: a complex multi-stage task involving opening a drawer, picking and placing a pen inside, and closing the drawer; 
(4) \textit{Deformable Object Manipulation}: folding a fabric towel. 
We collect an average of 100 demonstrations per task to facilitate embodiment alignment. Each model is evaluated over 20 trials per random seed, totaling 60 trials per task with diverse object placements. The success rate across these 60 trials is reported as the primary performance metric.

\begin{figure}
    \centering
    \includegraphics[width=1.0\linewidth]{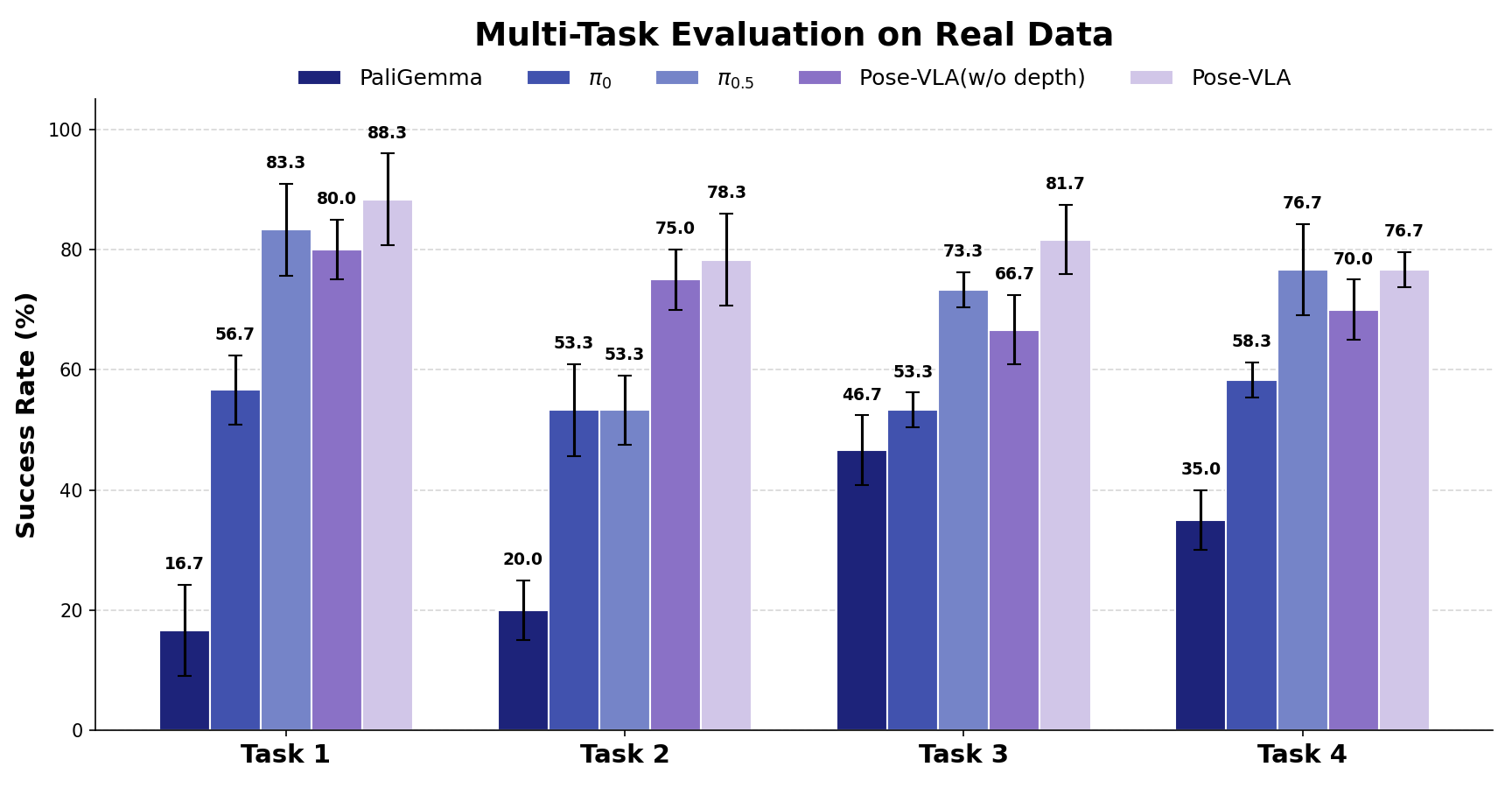}
       \vspace{-0.15in}
    \caption{Success rate comparison of Pose-VLA and baseline models across four real-world manipulation tasks. Each model is evaluated over 60 trials per task, with success rates reported as percentages. Under the same demonstration scale, Pose-VLA consistently outperforms current vision-language-action baselines, especially in long-horizon and deformable object tasks. \label{fig:hist} }
   \vspace{-0.1in}
\end{figure}

\noindent\textbf{Results and Analysis.}
As shown in Fig~\ref{fig:hist}, Pose-VLA demonstrates superior manipulation capability with an average success rate of $81.25\%$, significantly outperforming the vanilla PaliGemma ($29.6\%$) and the strong baseline $\pi_{0.5}$ ($71.65\%$). Our method consistently outperforms or matches $\pi_{0.5}$ across all tasks, notably achieving significant improvements in Task 2 ($78.3\%$ vs. $53.3\%$) and Task 3 ($81.7\%$ vs. $73.3\%$). Furthermore, the ablation study highlights the critical role of the depth modality; removing depth information leads to a marked performance decline, particularly in Task 3 where the success rate drops by $15.0$ percentage points (from $81.7\%$ to $66.7\%$). This substantial gap, alongside the $8.3\%$ drop in Task 1 (from $88.3\%$ to $80.0\%$), confirms that explicit 3D geometry is indispensable for precise spatial alignment and minimizing execution ambiguity.

\subsection{Ablation Study}
To verify how our 3D spatial pre-training and key design choices facilitate downstream performance, we conduct a series of ablation studies on both the 3D perception and RoboTwin 2.0 benchmarks. We systematically evaluate the impact of input modalities, pre-training objectives, architectural backbones, and action representations.

\noindent\textbf{Ablation of Geometric Modalities.}
We first investigate the contribution of each geometric modality on 3D grounding tasks, as shown in Table~\ref{tab:ablation_3d}. On the SUN RGB-D benchmark, which contains complex indoor layouts, the integration of depth information provides a pivotal boost, increasing the $\text{AP}_{15}$ from 39.0 (\textit{RGB Only}) to 45.1 (\textit{RGB+Depth}). In contrast, the performance gain on the Objectron dataset is less pronounced. We attribute this to the nature of the Objectron data, which primarily provides sparse and localized depth points centered around specific objects, offering limited additional geometric context compared to the dense depth maps in SUN RGB-D. Notably, the combination of Raymap and Depth (\textit{Full Model}) yields the best overall performance, confirming the synergy between explicit ray-based priors and metric depth.

\begin{table}[h]
\centering
\renewcommand{\arraystretch}{1.1}
\caption{Ablation study of geometric modalities on 3D Grounding.}\label{tab:ablation_3d}
\footnotesize
\begin{tabular}{lcc}
\toprule
Configuration & SUN RGB-D~\cite{song2015sun} & Objectron~\cite{ahmadyan2021objectron} \\ 
\midrule
RGB Only & 39.0 & 86.4 \\
RGB + Raymap & 38.5 & 86.7 \\
RGB + Depth & 45.1 & 85.4 \\
\textbf{Full Model} & \textbf{45.5} & \textbf{87.3} \\
\bottomrule
\end{tabular}
\end{table}

To thoroughly evaluate our design choices for robotic tasks, we perform two sets of ablation studies on the RoboTwin 2.0 benchmark. First, in Table~\ref{tab:incremental_ablation}, we present an \textit{incremental analysis} to trace how each component, ranging from data scaling to action space optimization, contributes to the final performance of Pose-VLA. Second, in Table~\ref{tab:comprehensive_ablation}, we conduct an \textit{orthogonal comparison} to validate specific architectural choices, such as the VLM backbone and tokenization strategy, against prominent alternatives. Crucially, to eliminate confounding variables and ensure a strictly fair comparison, every evaluated variant is fine-tuned under a unified regime of 80K optimization steps with a batch size of 32, starting directly from its original pre-trained checkpoint.

\begin{table}[htb]
\vspace{-4mm}
\centering
\footnotesize 
\setlength{\tabcolsep}{4pt} 
\caption{\textbf{Incremental} ablation study on the RoboTwin 2.0 benchmark. \textcolor{green}{Green} and \textcolor{red}{Red} denote relative performance changes from the preceding row. \protect(80K steps, and batch size=32).}
\label{tab:incremental_ablation}
\begin{tabular}{ll c@{\,\,}l c@{\,\,}l} 
\toprule
\multirow{2}{*}{\textbf{Type}} & \multirow{2}{*}{\textbf{Improvements}} & \multicolumn{4}{c}{\textbf{Success Rate (\%)}} \\
\cmidrule(l){3-6} 
& & \textbf{Easy} & & \textbf{Hard} & \\ 
\midrule
Baseline & PaliGemma + Action-Expert & 35.4 & & 33.4 & \\
\midrule
\multirow{2}{*}{Data} & + 3D Data & 70.2 & \textcolor{green}{\tiny (+34.6)} & 69.1 & \textcolor{green}{\tiny (+35.7)} \\
& + 3D Data \& Robotic Data & 72.9 & \textcolor{green}{\tiny (+2.7)} & 72.0 & \textcolor{green}{\tiny (+2.9)} \\
\midrule 
\multirow{2}{*}{Action} & Post-train: Joint $\rightarrow$ EE pose & \textbf{79.9} & \textcolor{green}{\tiny (+7.0)} & \textbf{79.1} & \textcolor{green}{\tiny (+7.1)} \\ 
& Pre-train: Cam $\rightarrow$ Base & 80.4 & \textcolor{green}{\tiny (+0.5)} & 76.7 & \textcolor{red}{\tiny (-2.4)} \\
\bottomrule
\end{tabular}
\vspace{-2mm}
\end{table}
\noindent\textbf{Ablation of Pose Pre-training Data.} 
We evaluate our data scaling strategy and mixture compositions across both Table~\ref{tab:incremental_ablation} and Table~\ref{tab:comprehensive_ablation}. As shown in the incremental analysis (Table~\ref{tab:incremental_ablation}), incorporating diverse non-robotic 3D grounding data to the vanilla \textit{PaliGemma} baseline yields a substantial $+35.7\%$ boost in success rate. This confirms that pre-training on 3D spatial tasks establishes a more robust geometric foundation than conventional VQA-dominated pre-training. To further isolate the impact of these data sources, Table~\ref{tab:comprehensive_ablation} compares orthogonal data mixtures. Crucially, pre-training on robot data only severely degrades results, proving that generic 3D vision data is the key driver of our framework's spatial capabilities. Conversely, relying solely on non-robotic 3D data (\textit{3D data only}) maintains highly competitive performance. Together, these results demonstrate that a strong grasp of 3D geometry constitutes the most significant contribution to success in our setting, effectively enhancing data efficiency by alleviating the reliance on costly, large-scale robotic demonstrations.

\noindent\textbf{Ablation of Action Representation.} 
As shown in Table~\ref{tab:incremental_ablation}, shifting the action space from joint angles to end-effector (EE) pose yields an average performance gain of over $+7.0\%$.
This improvement stems from the fact that our VLM is pre-trained to reason over 3D spatial poses; thus, adopting an EE pose interface allows the model to more directly leverage its learned geometric features. In contrast, joint-space control introduces a transformation overhead that degrades knowledge transfer. Conversely, we observe that predicting actions in the base frame leads to a performance drop ($-2.4\%$) in \textit{RoboTwin-Hard} scenarios. This indicates that camera-centric representations provide more robustness when facing significant visual variations in downstream policy transfer.

\begin{table}[htb]
\vspace{-4mm}
\centering
\small
\caption{\textbf{Orthogonal} ablation study on the RoboTwin 2.0 benchmark. SR denotes success rate. \protect(80K steps, and batch size=32).}
\label{tab:comprehensive_ablation}
\setlength{\tabcolsep}{5pt}
\renewcommand{\arraystretch}{1.1}
\resizebox{\columnwidth}{!}{%
\begin{tabular}{llcc}
\toprule
\textbf{Dimension} & \textbf{Variant} & \textbf{Easy SR (\%)} & \textbf{Hard SR (\%)} \\
\midrule
Backbone & Qwen3PI & 57.7 & 55.3 \\
\midrule
Tokenization & Pose-VLA (OpenVLA binning) & 68.5 & 65.9 \\
\midrule
\multirow{2}{*}{Data Mixture} & Pose-VLA (Robotic data only) & 71.3 & 68.3 \\
 & Pose-VLA (3D data only) & 77.2 & 76.2 \\
\midrule
\textbf{Full Model} & \textbf{Pose-VLA} & \textbf{79.9} & \textbf{79.1} \\
\bottomrule
\end{tabular}%
}
\vspace{-2mm}
\end{table}
\noindent\textbf{Ablation of VLM Backbone.} 
To verify whether the performance gain of Pose-VLA stems from our specific pre-training rather than simply the scale of the VLM backbone, we evaluate a baseline using Qwen3-VL~\cite{qwen3vl2025} (denoted as \textit{Qwen3PI}). 
Although Qwen3-VL possesses state-of-the-art 3D grounding capabilities, Table~\ref{tab:comprehensive_ablation} shows that Qwen3PI, which undergoes only post-training, trails Pose-VLA by more than 20\%, despite outperforming the scratch baseline.
This gap confirms that even powerful VLMs lack robust 3D representations necessary for fine-grained manipulation without explicit pose-centric pre-training. Furthermore, our approach shows superior data efficiency under \textit{tighter demonstration budgets}. Remarkably, this advantage persists even when pre-training utilizes \textit{solely non-robotic 3D data}, validating pose tokens as a ``universal interface'' to transfer geometric knowledge to robotic policies.


\noindent\textbf{Ablation of Tokenization.} 
We evaluate the structural advantages of our pose token representation by comparing it against standard methods, as reported in Table~\ref{tab:comprehensive_ablation}. Specifically, we contrast our approach with the uniform binning of delta actions used in OpenVLA~\cite{kim2024openvla}. OpenVLA's tokenization is fundamentally limited in two ways: \textbf{1)} It relies on \textit{local delta} representations tied to specific robot kinematics, which inherently blocks the absorption of non-robotic data. In contrast, our \textit{global spatial} representation decouples actions from specific embodiments, serving as a ``universal interface'' that seamlessly integrates both robotic and non-robotic 3D trajectories. \textbf{2)} OpenVLA employs \textit{uniform binning}, which inefficiently utilizes token capacity. Our \textit{non-uniform} approach, detailed in the Appendix, allocates higher resolution at close proximities, yielding significantly higher precision for delicate, near-contact manipulation. Consequently, reverting to OpenVLA's standard tokenization results in strictly inferior performance, validating that our unique design is essential for both data scaling and precise 3D manipulation.

\section{Limitations and Future Work}
\label{sec:limitation}
While \textit{Pose-VLA} provides a robust spatial foundation for robotic manipulation, its current scope presents promising avenues for future exploration. First, due to computational constraints, the full potential of large-scale 3D data scaling and high-efficiency 3D annotation for robotic videos remains partially unexplored. Furthermore, while our framework inherently supports deformable object manipulation during downstream adaptation, such as towel folding, the pre-training stage is currently driven by rigid-object datasets. Explicitly formulating deformable states, such as 3D keypoints or dense grasp annotations, into the pose token vocabulary during the large-scale pre-training phase represents a critical next step. Alongside this, extending these spatial priors from parallel-jaw grippers to multi-fingered dexterous hands will further unlock complex manipulation capabilities. Additionally, scaling our real-world experiments to a broader diversity of behaviors in unstructured environments remains a key objective. In particular, transitioning from static tabletop setups to mobile and egocentric manipulation presents an exciting frontier for future work. Finally, as our core focus in this work is maximizing spatial priors, we observe a natural trade-off regarding the backbone's original semantic reasoning when facing complex semantic distractors. Exploring semantic-preserving training paradigms to maintain general reasoning capabilities without compromising 3D awareness offers a compelling direction for future research.

\section{Conclusion}
We present Pose-VLA, a framework utilizing a unified pose token as a universal interface to bridge spatial perception and robotic control. By integrating RGB-D data with camera intrinsics, our architecture instills an intrinsic 3D awareness that enables the learning of robust spatial and motion priors through large-scale pre-training. Pose-VLA achieves state-of-the-art results in 3D grounding and provides a generalizable physical foundation for downstream robotic tasks, as validated across LIBERO, RoboTwin, and real-world experiments. 
Our work demonstrates a viable avenue for scaling VLA models by leveraging heterogeneous non-robotic 3D data alongside specialized robotic demonstrations. We advocate for a shift from VQA-based foundations toward pre-training embodied-aware VLMs, fostering the development of backbones that are inherently grounded in the physical world.

\bibliographystyle{plainnat}
\bibliography{egbib}

\clearpage

\appendix
This supplemental material is organized as follows: 
In Section~\ref{sec:discretization}, we provide a detailed analysis of our distribution-aware pose discretization strategy and the initialization of the spatial tokens. 
Section~\ref{sec:robotwin_detail} presents comprehensive success rates across all 50 tasks in the RoboTwin 2.0 benchmark, alongside an ablation study on action expert pre-training. 
In Section~\ref{sec:tsne}, we investigate the latent representations via t-SNE visualizations to demonstrate the discriminability of {Pose-VLA} features. 
Finally, we showcase additional qualitative results and real-world robotic executions in the \textbf{accompanying video}.

\subsection{Distribution-Aware Pose Discretization}
\label{sec:discretization}

To integrate spatial grounding into the VLM's expanded vocabulary, we discretize continuous 6-DoF poses into a sequence of discrete tokens. A {naive uniform discretization} across the workspace often leads to a sub-optimal allocation of representational capacity, as it overlooks the {heteroscedastic precision requirements} inherent in robotic manipulation. Specifically, near-field interactions demand higher control resolution, whereas far-field movements are naturally more tolerant of spatial quantization errors.

\begin{figure}[h]
    \centering
    \includegraphics[width=0.95\linewidth]{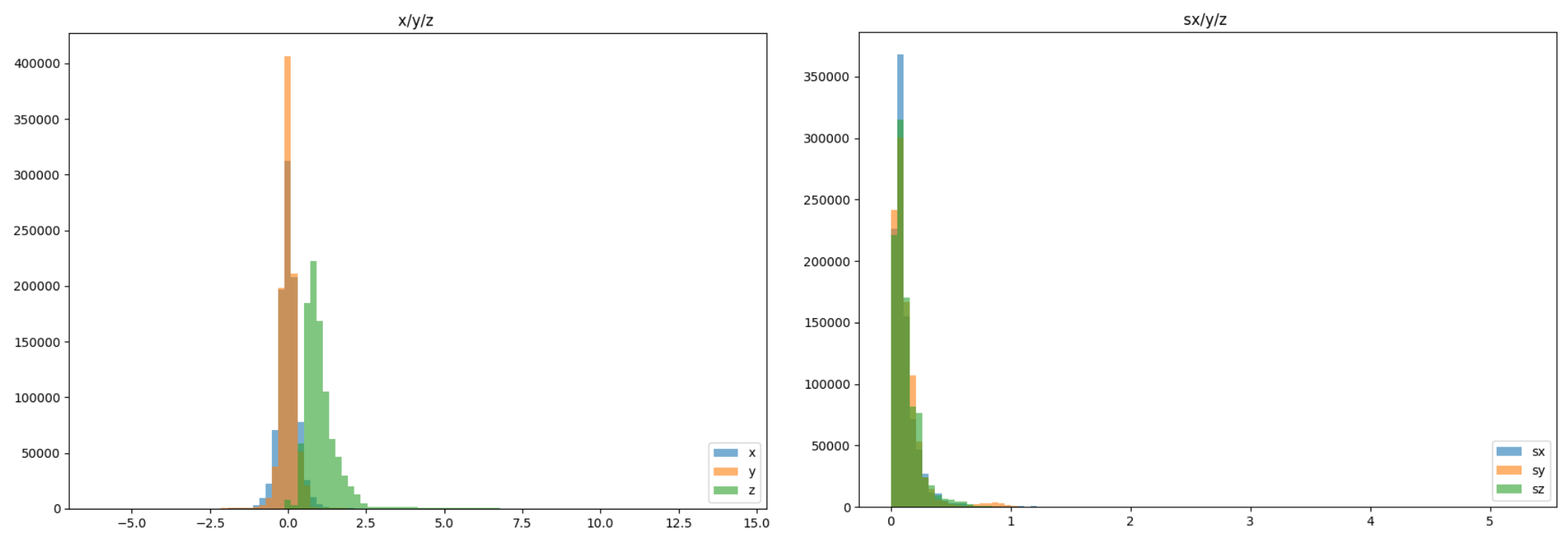}
       \vspace{-0.1in}
    \caption{Data statistics of object translation and size in datasets. Translations along the x and y axes show similar distributions, while the z-axis translation exhibits a notably different pattern. In contrast, object sizes are similarly distributed across all three axes. \label{fig:distribution} }
   \vspace{-0.1in}
\end{figure}

An analysis of our training statistics reveals a distinct distributional disparity between the horizontal ($x, y$) and vertical ($z$) axes, as illustrated in Fig.~\ref{fig:distribution}. Specifically, the $x$ and $y$ coordinates are densely concentrated around the origin. Conversely, the $z$ axis exhibits a highly skewed distribution, characterized by a positive offset and a extended right tail.


To address this, we adopt a {non-uniform discretization strategy}. Specifically, we compute bin boundaries such that the bin width is inversely proportional to the data density, resulting in narrower bins within high-density regions and wider bins at the distribution tails. This ensures that the resulting pose tokens are utilized with {approximately equal frequency} across the training set, thereby maximizing the {informational entropy} of the discrete representation. Following this approach, we extend the PaliGemma vocabulary with specialized tokens $\texttt{<rot>}$, $\texttt{<trans\_xy>}$, $\texttt{<trans\_z>}$, and $\texttt{<size>}$, assigning $N = 2048$ bins to each. While alternative quantization methods could potentially offer further refinements, they fall outside the primary scope of this work and are reserved for future exploration.

\subsection{Detailed Results on RoboTwin}
\label{sec:robotwin_detail}
Table~\ref{tab:robotwin-short} presents the individual success rates for all 50 tasks within the RoboTwin 2.0~\cite{chen2025robotwin} benchmark. Furthermore, we conduct an ablation study regarding the \textbf{pre-training of the action expert} to evaluate its impact on downstream performance. We compare our full model, {Pose-VLA} ($\pi_0$), which initializes the action expert from official $\pi_0$ checkpoints, against {Pose-VLA} (scratch), where the action expert is trained from randomly initialized weights. The results indicate that while action pre-training provides benefits, the spatial pre-training on the VLM backbone contributes most significantly to the overall performance gains.

\begin{table*}[ht]
  \centering
  \footnotesize
  \setlength{\tabcolsep}{6.5pt}
  \caption{\textbf{Comparison of our method and VLA models on RoboTwin 2.0 simulation}}
  \label{tab:robotwin-short}
  \begin{tabular}{*{1}{>{\centering\arraybackslash}m{3.2cm}} *{10}{>{\centering\arraybackslash}m{0.75cm}}}
    \toprule
    \multirow{2}{*}{\textbf{Manipulation Task}} 
      & \multicolumn{2}{c}{$\mathbf{\pi}_0$}
      & \multicolumn{2}{c}{$\mathbf{\pi}_{0.5}$}
      & \multicolumn{2}{c}{PaliGemma\_expert}
      & \multicolumn{2}{c}{\textit{Pose-VLA}} 
      & \multicolumn{2}{c}{\textit{Pose-VLA ($\boldsymbol{\pi}_0$)}} \\
    & Easy & Hard & Easy & Hard & Easy & Hard & Easy & Hard & Easy & Hard\\ 
    \midrule
\textit{Adjust Bottle} & 70\% & 57\% & 62\% & 69\% & 67\% & 47\% & 97\% & 77\% & 97\% & 90\% \\
\textit{Beat Block Hammer} & 80\% & 73\% & 78\% & 93\% & 40\% & 37\% & 100\% & 87\% & 87\% & 87\% \\
\textit{Blocks Ranking Rgb} & 80\% & 83\% & 86\% & 85\% & 47\% & 40\% & 83\% & 77\% & 83\% & 80\% \\
\textit{Blocks Ranking Size} & 43\% & 60\% & 76\% & 26\% & 17\% & 3\% & 60\% & 73\% & 80\% & 60\% \\
\textit{Click Alarmclock} & 83\% & 73\% & 92\% & 89\% & 80\% & 57\% & 83\% & 93\% & 87\% & 90\% \\
\textit{Click Bell} & 67\% & 83\% & 98\% & 66\% & 57\% & 57\% & 87\% & 83\% & 97\% & 97\% \\
\textit{Dump Bin Bigbin} & 100\% & 90\% & 89\% & 97\% & 73\% & 77\% & 97\% & 97\% & 97\% & 97\% \\
\textit{Grab Roller} & 93\% & 97\% & 100\% & 100\% & 53\% & 57\% & 97\% & 100\% & 100\% & 100\% \\
\textit{Handover Block} & 33\% & 40\% & 84\% & 57\% & 7\% & 17\% & 73\% & 80\% & 70\% & 60\% \\
\textit{Handover Mic} & 90\% & 100\% & 92\% & 97\% & 83\% & 87\% & 93\% & 93\% & 100\% & 100\% \\
\textit{Hanging Mug} & 20\% & 21\% & 22\% & 17\% & 7\% & 10\% & 27\% & 23\% & 13\% & 33\% \\
\textit{Lift Pot} & 63\% & 73\% & 100\% & 85\% & 37\% & 30\% & 100\% & 97\% & 93\% & 98\% \\
\textit{Move Can Pot} & 57\% & 53\% & 72\% & 55\% & 23\% & 20\% & 63\% & 53\% & 60\% & 70\% \\
\textit{Move Pillbottle Pad} & 73\% & 67\% & 85\% & 61\% & 10\% & 20\% & 90\% & 87\% & 93\% & 86\% \\
\textit{Move Playingcard Away} & 80\% & 83\% & 96\% & 84\% & 33\% & 30\% & 87\% & 83\% & 100\% & 96\% \\
\textit{Move StaplerPad} & 37\% & 50\% & 61\% & 42\% & 7\% & 10\% & 70\% & 53\% & 73\% & 72\% \\
\textit{Open Laptop} & 73\% & 70\% & 88\% & 96\% & 67\% & 70\% & 93\% & 93\% & 93\% & 98\% \\
\textit{Open Microwave} & 71\% & 83\% & 71\% & 77\% & 70\% & 83\% & 60\% & 63\% & 60\% & 82\% \\
\textit{Pick Diverse Bottles} & 40\% & 53\% & 20\% & 71\% & 7\% & 3\% & 60\% & 77\% & 53\% & 50\% \\
\textit{Pick Dual Bottles} & 47\% & 50\% & 21\% & 63\% & 10\% & 0\% & 87\% & 87\% & 77\% & 54\% \\
\textit{Place A2b Left} & 60\% & 43\% & 84\% & 82\% & 27\% & 10\% & 90\% & 80\% & 93\% & 86\% \\
\textit{Place A2b Right} & 30\% & 47\% & 83\% & 84\% & 13\% & 7\% & 77\% & 80\% & 80\% & 88\% \\
\textit{Place Bread Basket} & 77\% & 47\% & 83\% & 64\% & 30\% & 17\% & 83\% & 73\% & 80\% & 70\% \\ 
\textit{PlaceBread Skillet} & 63\% & 60\% & 86\% & 66\% & 3\% & 3\% & 93\% & 77\% & 77\% & 74\% \\
\textit{Place Burger Fries} & 93\% & 93\% & 97\% & 87\% & 17\% & 27\% & 83\% & 83\% & 90\% & 82\% \\ 
\textit{Place Can Basket} & 45\% & 40\% & 53\% & 62\% & 27\% & 23\% & 53\% & 70\% & 87\% & 72\% \\ 
\textit{Place Cans Plasticbox} & 83\% & 73\% & 90\% & 84\% & 23\% & 17\% & 97\% & 87\% & 97\% & 90\% \\ 
\textit{Place Container Plate} & 93\% & 100\% & 89\% & 95\% & 90\% & 80\% & 97\% & 100\% & 97\% & 98\% \\ 
\textit{Place Dual Shoes} & 63\% & 50\% & 93\% & 75\% & 3\% & 0\% & 87\% & 77\% & 80\% & 82\% \\ 
\textit{Place Empty Cup} & 100\% & 97\% & 96\% & 99\% & 67\% & 77\% & 100\% & 97\% & 100\% & 100\% \\ 
\textit{Place Fan} & 57\% & 73\% & 87\% & 85\% & 13\% & 10\% & 77\% & 77\% & 83\% & 82\% \\ 
\textit{Place Mouse Pad} & 43\% & 40\% & 63\% & 39\% & 13\% & 13\% & 43\% & 57\% & 73\% & 66\% \\ 
\textit{Place Object Basket} & 60\% & 73\% & 60\% & 76\% & 20\% & 20\% & 77\% & 80\% & 90\% & 86\% \\
\textit{Place Object Scale} & 70\% & 43\% & 82\% & 80\% & 10\% & 10\% & 67\% & 83\% & 77\% & 74\% \\ 
\textit{Place Object Stand} & 70\% & 77\% & 90\% & 85\% & 27\% & 43\% & 83\% & 80\% & 87\% & 82\% \\ 
\textit{Place Phone Stand} & 70\% & 57\% & 83\% & 81\% & 30\% & 20\% & 87\% & 87\% & 83\% & 84\% \\ 
\textit{Place Shoe} & 87\% & 80\% & 96\% & 93\% & 23\% & 30\% & 100\% & 87\% & 97\% & 96\% \\ 
\textit{Press Stapler} & 70\% & 80\% & 80\% & 83\% & 70\% & 67\% & 83\% & 83\% & 83\% & 84\% \\ 
\textit{Put Bottles Dustbin} & 63\% & 83\% & 76\% & 79\% & 13\% & 13\% & 80\% & 81\% & 90\% & 84\% \\ 
\textit{Put Object Cabinet} & 53\% & 43\% & 83\% & 79\% & 26\% & 7\% & 62\% & 63\% & 63\% & 71\% \\ 
\textit{Rotate Qrcode} & 67\% & 60\% & 79\% & 87\% & 27\% & 23\% & 70\% & 70\% & 67\% & 76\% \\ 
\textit{Scan Object} & 47\% & 23\% & 80\% & 65\% & 7\% & 3\% & 87\% & 90\% & 70\% & 62\% \\ 
\textit{Shake Bottle Horizontally} & 100\% & 90\% & 100\% & 99\% & 87\% & 83\% & 100\% & 100\% & 100\% & 94\% \\ 
\textit{Shake Bottle} & 97\% & 90\% & 97\% & 97\% & 70\% & 73\% & 100\% & 100\% & 100\% & 94\% \\ 
\textit{Stack Blocks Three} & 50\% & 63\% & 82\% & 76\% & 11\% & 7\% & 47\% & 44\% & 77\% & 84\% \\ 
\textit{Stack Blocks Two} & 80\% & 93\% & 100\% & 100\% & 27\% & 27\% & 93\% & 90\% & 97\% & 94\% \\ 
\textit{Stack Bowls Three} & 77\% & 63\% & 77\% & 71\% & 39\% & 63\% & 69\% & 73\% & 60\% & 88\% \\
\textit{Stack Bowls Two} & 93\% & 97\% & 93\% & 96\% & 83\% & 90\% & 93\% & 100\% & 87\% & 96\% \\ 
\textit{Stamp Seal} & 60\% & 53\% & 73\% & 55\% & 46\% & 23\% & 67\% & 73\% & 60\% & 70\% \\ 
\textit{Turn Switch} & 27\% & 27\% & 57\% & 54\% & 33\% & 27\% & 43\% & 37\% & 26\% & 27\% \\
    \midrule
    \textbf{\textit{Average (\%)}} & 67.00 & 65.12  & 79.48 & 76.16 & 35.40 & 33.36 & 79.91 & 79.10 & \textbf{81.30} & \textbf{80.72} \\
    \bottomrule
  \end{tabular}
\end{table*}

\subsection{T-SNE Visualization of VLM Features}
\label{sec:tsne}
\begin{figure}[t]
    \centering
    \includegraphics[width=1.0\linewidth]{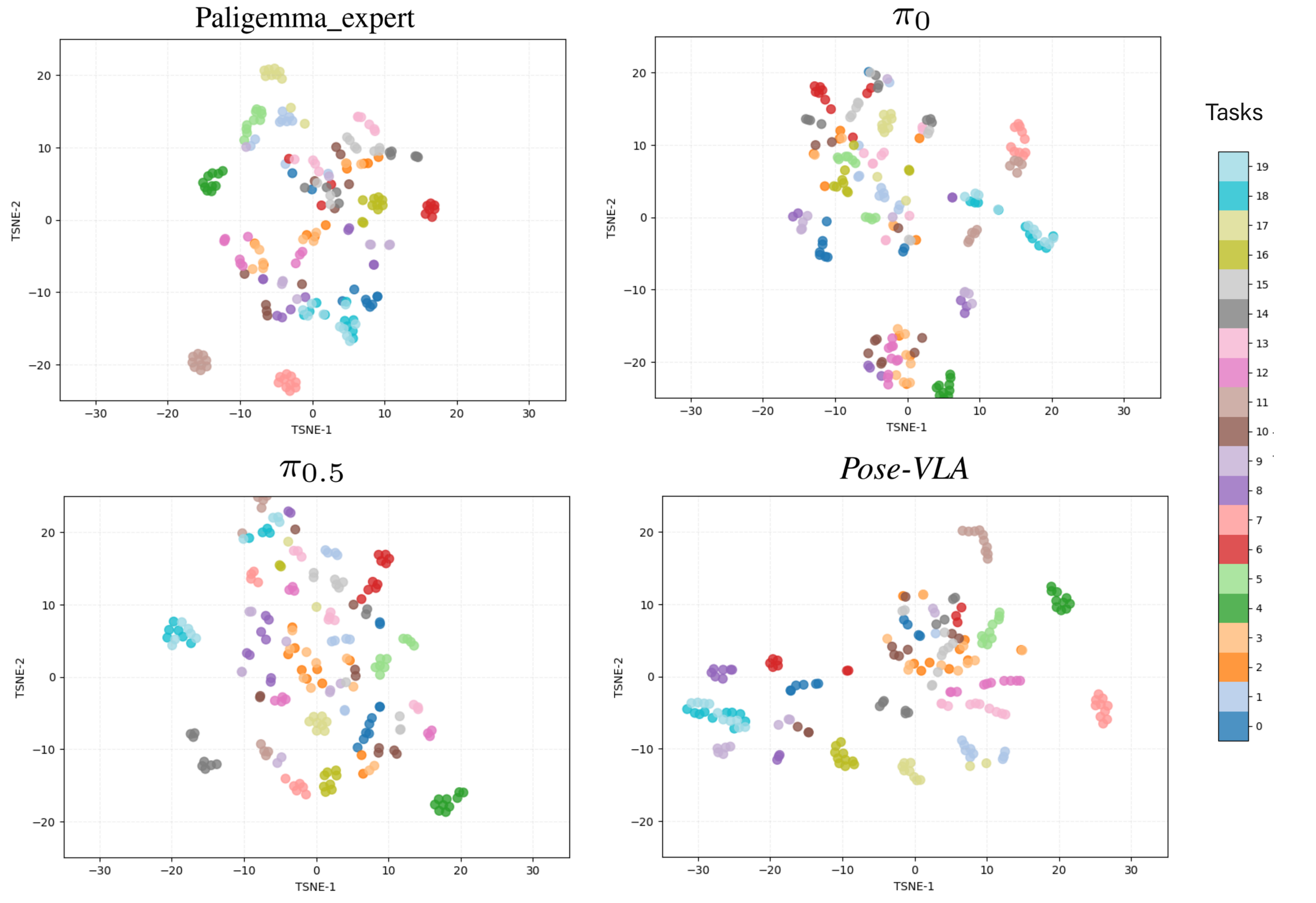}
    \vspace{-0.15in}
    \caption{T-SNE visualization of VL features across 20 tasks in RoboTwin 2.0. Each point represents a feature vector at a specific timestep, with colors denoting different tasks. Our {Pose-VLA} exhibits clearer inter-task separation and tighter intra-task clusters compared to $\pi_0$ and $\pi_{0.5}$, effectively mitigating the representation collapse observed in baseline methods.}
    \label{fig:feature}
    \vspace{-0.1in}
\end{figure}

To further investigate the latent representations, we visualize the Vision-Language (VL) features from various VLA methods using t-SNE. As illustrated in Fig.~\ref{fig:feature}, we compare the token embeddings of \textit{PaliGemma-with-Expert}, $\pi_0$, $\pi_{0.5}$, and our {Pose-VLA}. 

While \textit{Pose-VLA} maintains well-separated clusters for target tasks, \textit{PaliGemma-with-Expert} exhibits significant overlap across classes. This indicates that vanilla action fine-tuning often leads to representation collapse, where the model loses the ability to distinguish between diverse task semantics. In contrast, \textit{Pose-VLA} features demonstrate superior inter-task discriminability and intra-task consistency, forming more compact and well-defined clusters that facilitate more robust policy learning.

\end{document}